\documentclass[runningheads]{llncs}
\usepackage[T1]{fontenc}
\usepackage{graphicx}
\usepackage{booktabs}
\usepackage{multirow}
\usepackage{tabularx}

\usepackage{tikz}
\usetikzlibrary{positioning, calc}
\usepackage{hyperref}

\begin{document}
\title{Teaching LLMs Brazilian Healthcare: Injecting Knowledge from Official Clinical Guidelines}
\titlerunning{Teaching LLMs Brazilian Healthcare}
%
%
%
%


\author{Hugo Abonizio\inst{1,2} \and
Filipe Rocha Lopes\inst{3} \and
Roberto Lotufo\inst{1} \and
Rodrigo Nogueira\inst{2}}

\authorrunning{Abonizio et al.}

\institute{FEEC, UNICAMP, Campinas, Brazil \and
Maritaca AI, Brazil \and
NoHarm.ai, Brazil}

\maketitle              

\setcounter{footnote}{0}

\begin{abstract}
Brazil's Unified Health System (SUS) relies on official clinical guidelines that define diagnostic criteria, treatments, dosages, and monitoring procedures for over 200 million citizens. Yet current LLMs perform poorly on this guideline-specific knowledge, and no benchmark evaluates clinical recall grounded in Brazilian Portuguese protocols. We address this gap by adapting Qwen2.5-14B-Instruct to the Brazilian clinical domain. From 178 official guidelines (${\sim}$5.4M tokens), we generate ${\sim}$70M tokens of synthetic data in three formats---rephrases, wiki-style articles, and question-answer pairs---using four generator LLMs. We then apply continual pre-training followed by Group Relative Policy Optimization (GRPO). We introduce HealthBench-BR, with 1,780 balanced true/false clinical assertions, and PCDT-QA, with 890 open-ended clinical questions scored by an LLM judge. Our best model achieves 83.9\% on HealthBench-BR and 85.4\% on PCDT-QA, outperforming GPT-5.2, Claude Sonnet 4.6, Gemini 3.1 Pro, and Google AI Overview's web-grounded RAG despite having only 14B parameters. Ablations show that generator diversity and reinforcement learning are critical to these gains. We release all datasets, benchmarks, and model weights to support reproducible clinical NLP research for Brazilian Portuguese. Code, data, and model weights are available at \url{https://github.com/hugoabonizio/clinical-protocols-br}.

\end{abstract}
\section{Introduction}

Brazil's Unified Health System (SUS -- \textit{Sistema Único de Saúde}) provides universal healthcare coverage to over 200 million people, making it one of the largest public health systems in the world. Clinical decision-making within SUS is governed by official clinical guidelines published by the Ministry of Health, which prescribe diagnostic criteria, treatment regimens, dosages, and monitoring procedures for conditions treated under public care~\cite{lei12401}.
These documents---primarily Clinical Protocols and Therapeutic Guidelines (PCDTs), but also several related categories described in Section~\ref{sec:corpus}---are hosted under the Ministry's PCDT portal.\footnote{\url{https://www.gov.br/saude/pt-br/assuntos/pcdt}}
Adherence to these protocols is mandatory for SUS managers, yet their volume and complexity make it difficult for healthcare professionals to recall specific recommendations at the point of care. Large language models (LLMs) offer a promising path toward intelligent clinical decision support, yet current medical LLMs are predominantly trained on English-language medical literature and evaluated on English benchmarks, leaving their knowledge of Brazilian clinical guidelines largely untested~\cite{brunetiseverino2025revalida,garcia2025stepforward}.

\begin{figure}[t]
\centering
\includegraphics[width=\columnwidth]{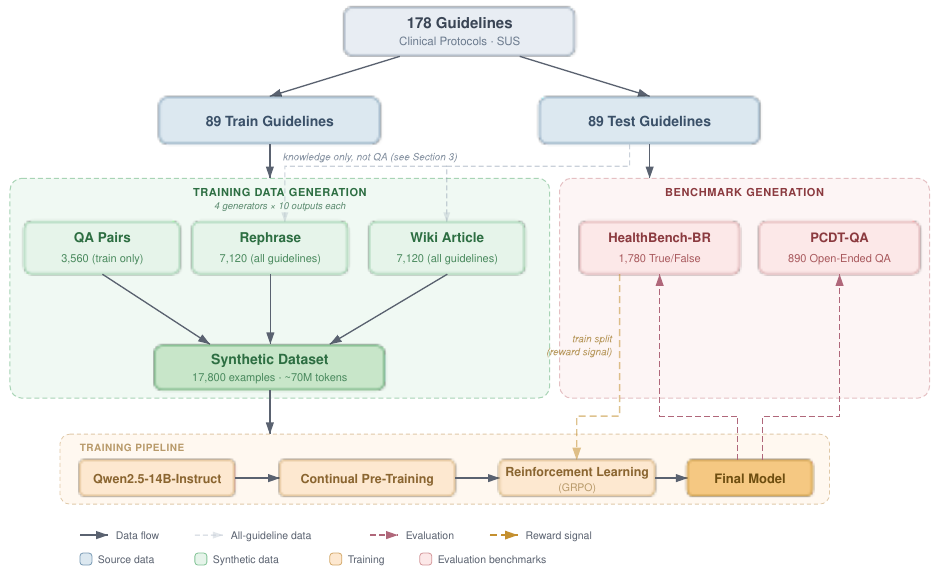}
\caption{Overview of our pipeline. The 178 clinical guidelines are split into train and test sets. All guidelines feed rephrase and wiki-style generation (dashed gray arrows), while QA pairs are generated only from train guidelines to prevent leakage. Four generator LLMs produce the synthetic corpus ({$\sim$}70M tokens), which is used for continual pre-training of Qwen2.5-14B-Instruct. The HealthBench-BR train split provides the reward signal for GRPO reinforcement learning (dashed amber arrow). Both benchmarks are evaluated on the final model (dashed red arrows).}
\label{fig:pipeline}
\end{figure}


Even state-of-the-art open medical LLMs leave Brazilian clinical knowledge largely uncovered. MedGemma~\cite{sellergren2025medgemma} is built on Anglo-centric data---biomedical text, FHIR-based electronic health records, and English QA pairs---with no multilingual coverage. BioMistral~\cite{labrak2024biomistral} is pre-trained exclusively on PubMed Central and assesses Portuguese only through machine-translated benchmarks. ApolloMoE~\cite{zheng2024apollomoe} covers 50 languages including Portuguese, but draws its medical corpus from generic multilingual sources rather than country-specific clinical protocols.

We address this gap by proposing a training protocol for adapting a general-purpose LLM to the clinical domain of the Brazilian public health system. Starting from all 178 clinical guidelines described above, we first generate approximately 70 million tokens of synthetic training data across three complementary formats (rephrases, wiki-style articles, and question-answer pairs) using four diverse generator LLMs, following multi-style rephrasing strategies shown to be effective for domain adaptation~\cite{maini2024wrap,cheng2024adaptllm}. This corpus is used for continual pre-training (CPT) of Qwen2.5-14B-Instruct~\cite{qwen25}, followed by reinforcement learning via Group Relative Policy Optimization (GRPO)~\cite{deepseek-math} to encourage explicit clinical reasoning. To evaluate knowledge absorption, we introduce two guideline-grounded benchmarks: \textbf{HealthBench-BR}, comprising 1{,}780 balanced true/false clinical assertions requiring precise factual recall, and \textbf{PCDT-QA}, containing 890 open-ended clinical questions scored by an LLM judge.

Our best configuration outperforms GPT-5.2 on both benchmarks despite being a 14B-parameter open model. In summary, our main contributions are: (i) a reproducible training protocol combining multi-generator synthetic data, continual pre-training, and reinforcement learning for clinical domain adaptation; (ii) two protocol-grounded evaluation benchmarks for the Brazilian public health system; (iii) a detailed analysis of the trade-offs between knowledge injection and catastrophic forgetting, including ablations showing that both multi-generator diversity and reinforcement learning are critical to the observed gains; and (iv) the public release of all synthetic datasets, benchmarks, and model weights to support reproducible research on clinical NLP for Brazilian Portuguese.

\section{Related Work}

\subsection{Medical LLMs and Portuguese Medical NLP}

Proprietary models have driven rapid progress on medical benchmarks, with Med-PaLM~\cite{singhal2023medpalm}, Med-PaLM~2~\cite{singhal2024medpalm2}, and Med-Gemini~\cite{saab2024medgemini} successively raising the bar on the USMLE, and GPT-4 exceeding the passing threshold by over 20 points without medical prompting~\cite{nori2023gpt4med}. Their closed nature and English-centric design, however, limit applicability in settings like Brazilian public health, where transparency and native-language reasoning are essential. On the open-source side, continual pre-training (CPT) has emerged as the dominant adaptation strategy: MEDITRON~\cite{chen2023meditron}, PMC-LLaMA~\cite{wu2024pmcllama}, Me-LLaMA~\cite{xie2024mellama}, and the more recent multimodal MedGemma~\cite{sellergren2025medgemma} scale CPT on biomedical literature from tens to over 100B tokens, while multilingual efforts---BioMistral~\cite{labrak2024biomistral} and ApolloMoE~\cite{zheng2024apollomoe}---either evaluate Portuguese only via machine translation or rely on generic multilingual corpora, with no incorporation of official Brazilian protocols.

Work specific to Portuguese medical NLP began with encoder models such as BioBERTpt~\cite{schneider2020biobertpt} for clinical NER, followed by Clinical-BR-LlaMA/Mistral~\cite{pinto2024clinicalbr} and DrBode~\cite{recogna2024drbode}, which applied LoRA and QLoRA on clinical notes and translated medical data---though adapted models often underperformed the base due to catastrophic forgetting. On the data and evaluation side, MedPT~\cite{farber2025medpt} released 384K native patient--doctor QA pairs, the Revalida exam has been used to benchmark LLMs~\cite{brunetiseverino2025revalida}, and HealthQA-BR~\cite{daddario2025healthqabr} provides 5,632 residency questions. None of these resources, however, evaluates factual recall over official Brazilian clinical protocols, and no prior work injects parametric knowledge from these protocols into LLMs---the dual gap we address with our adaptation pipeline and protocol-grounded benchmarks (Section~\ref{sec:benchmarks}).

\subsection{Synthetic Data for Domain Adaptation}

Training on high-quality synthetic data can match or exceed training on much larger natural corpora. Gunasekar et al.~\cite{gunasekar2023textbooks} demonstrated that models trained primarily on synthetic textbook-quality data achieve strong coding performance at a fraction of the usual data scale. WRAP~\cite{maini2024wrap} generalized this insight to pre-training: by prompting an instruction-tuned model to rephrase web documents in multiple styles---Wikipedia-like, question--answer, and paraphrase---it speeds up pre-training while improving downstream accuracy. Nemotron-CC~\cite{su2024nemotroncc} scaled this source-rephrasing approach to trillions of tokens, and AdaptLLM~\cite{cheng2024adaptllm} showed that reformatting raw domain corpora as reading comprehension tasks enables small models to rival much larger domain-specific ones. In medicine, AlpaCare~\cite{zhang2024alpacare} and ChatDoctor~\cite{li2023chatdoctor} demonstrated that LLM-generated instructions and dialogues substantially boost clinical performance from limited seeds. Our synthetic data pipeline follows this philosophy, transforming 178 official Brazilian protocols into rephrases, wiki-style articles, and QA pairs---similar to the multi-style strategy of WRAP but applied to a small, high-value domain corpus to maximize knowledge absorption.

\subsection{Catastrophic Forgetting in Continual Pre-Training}

A key risk in CPT is catastrophic forgetting, which scales as a power law in parameters and update steps~\cite{kalajdzievski2024scaling}. Parameter-efficient methods like LoRA naturally mitigate this by modifying fewer parameters~\cite{hu2022lora,biderman2024loraforgets}, and replay of even 1\% of pre-training data can prevent it~\cite{bethune2025scaling}---though recent work suggests that some observed drops reflect alignment loss rather than true knowledge erasure~\cite{zheng2025spurious}. We opt for full CPT to maximize knowledge injection from our small domain corpus, explicitly measure forgetting on general benchmarks, and provide ablations comparing the trade-offs of forgetting-mitigation strategies.

\section{Clinical Guideline Corpus}
\label{sec:corpus}




Brazilian clinical guidelines fall into several regulatory categories: Clinical Protocols and Therapeutic Guidelines (PCDTs -- \textit{Protocolos Clínicos e Diretrizes Terapêuticas}), which form the majority; Protocols of Use (\textit{Protocolos de Uso}); Oncology Diagnostic and Therapeutic Guidelines (\textit{Diretrizes Diagnósticas e Terapêuticas em Oncologia}); National Guidelines (\textit{Diretrizes Brasileiras}); and Care Pathways (\textit{Linhas de Cuidado}). All are elaborated with support from the National Commission for the Incorporation of Technologies (CONITEC -- \textit{Comissão Nacional de Incorporação de Tecnologias no SUS}), and their adoption is mandatory for SUS managers. For brevity, we refer to the full collection as \emph{clinical guidelines} throughout this paper.

We collected all 178 clinical guidelines available as of March 2026 as PDF documents from the Ministry of Health portal. The text content was extracted from each PDF, yielding a raw corpus of approximately 26.9 million characters. Individual protocols range from 18k to over 915k characters (median 117k). To fit within the context window of our generator models (Section~\ref{sec:synthesis}), we truncate each protocol at 120k characters, retaining the full text for the majority of protocols and the initial sections---which contain diagnostic criteria, treatment recommendations, and monitoring parameters---for longer ones. After truncation, the corpus comprises approximately 16.6M characters (5.4M tokens). The extracted protocols and all derived datasets are publicly available.



\section{Synthetic Data Pipeline}
\label{sec:synthesis}

Rather than training on the original guidelines, we use LLMs to transform each guideline into synthetic texts in three complementary formats (Figure~\ref{fig:pipeline}, green blocks), following recent work showing that rephrasing corpora into diverse formats improves knowledge absorption during pre-training~\cite{maini2024wrap,cheng2024adaptllm}. Each train-split guideline is processed under three prompts, while test-split guidelines are processed only under the rephrase and wiki-style prompts, sampling 10 outputs per prompt.

\paragraph{Rephrase.} The generator rewrites the guideline, modifying information order and paraphrasing while preserving all clinical content. This exposes the model to the same knowledge in varied surface forms. Rephrases are generated for all 178 guidelines.

\paragraph{Wiki-style article.} The generator produces an expository article in the style of Wikipedia, restructuring the prescriptive clinical document while retaining all key numerical values and details. Wiki articles are generated for all 178 guidelines.

\paragraph{Question--answer pairs.} The generator creates questions and detailed answers grounded in the guideline, ranging from broad clinical queries to specific dosage and monitoring questions. Answers include step-by-step rationales to encourage internalization of clinical reasoning patterns. To prevent data leakage, QA pairs are generated only from the 89 train-split guidelines, as illustrated by the dashed arrows in Figure~\ref{fig:pipeline}: test-split guidelines feed only rephrase and wiki generation, never QA.

To increase the stylistic and lexical diversity of the synthetic corpus, each of the three generation strategies is applied independently by four distinct models: GPT-4.1-mini, GPT-5-nano, GPT-OSS-20B, and Qwen3-235B. As shown in Table~\ref{tab:ablation-augtype}, scaling from one to four generators substantially improves downstream performance. We also evaluate self-generation with the target model 
(Qwen2.5-14B-Instruct), which yields the weakest results; we analyze this case in Section~\ref{sec:ablation-generators}.

The full synthetic corpus comprises 17{,}800 documents ({$\sim$}70M tokens), a roughly 13$\times$ expansion over the original 5.4M-token corpus with substantially greater format and stylistic diversity. This dataset is used for continual pre-training (Section~\ref{sec:setup}).

\section{Evaluation Benchmarks}
\label{sec:benchmarks}

\begin{figure}[t]
\centering
\includegraphics[width=\linewidth]{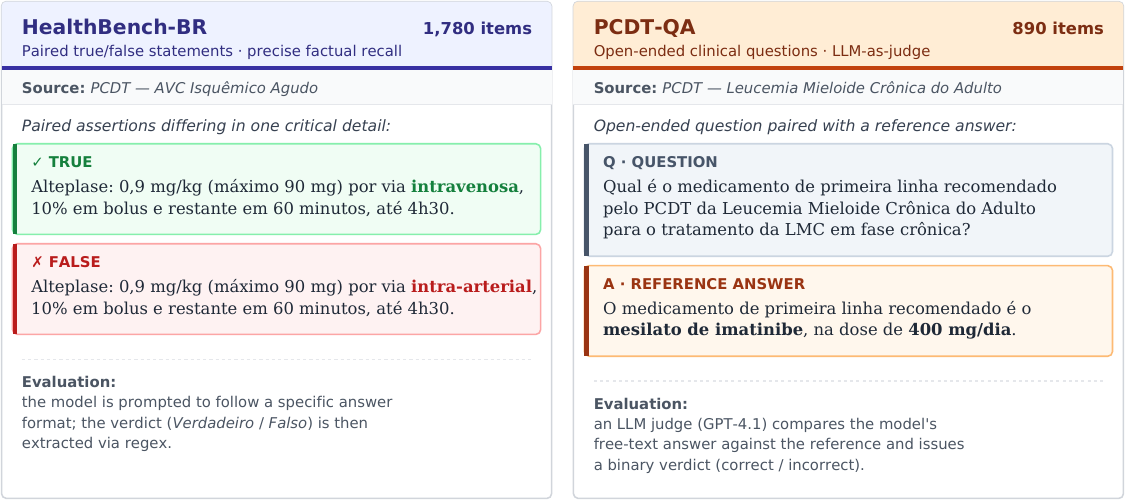}
\caption{Examples from the two proposed benchmarks. \textit{Left:} HealthBench-BR pairs each true assertion with a variant in which one critical detail is modified to produce a plausible distractor. \textit{Right:} PCDT-QA pairs an open-ended question with a reference answer, scored against the model's response by an LLM judge.}
\label{fig:benchmark_examples}
\end{figure}

We introduce two complementary benchmarks derived from the clinical guidelines, illustrated in Figure~\ref{fig:benchmark_examples}. Both are split into training and test by guideline, with 89 guidelines in each split. Questions are generated for all 178 guidelines, but we report accuracy separately on train and test splits; performance on test-split questions reflects generalization to guidelines seen only as rephrase and wiki data during training, never as QA supervision.

\subsection{HealthBench-BR}

HealthBench-BR is a true/false benchmark comprising 1{,}780 clinical assertions (10 per guideline, 890 per split). Assertions are generated in paired form using GPT-5.2, conditioned on the full text of each guideline: for each clinical fact, one statement is faithful to the guideline (true) and a second modifies a key detail---such as a dosage, route of administration, or monitoring interval---to produce a plausible but false variant. This yields a perfectly balanced dataset (50\% true, 50\% false) and ensures that correct classification requires precise factual recall rather than surface-level heuristics. During evaluation, the model is prompted to conclude with an explicit verdict (\textit{Verdadeiro}/\textit{Falso}), extracted via regex. All inference uses greedy decoding.

\subsection{PCDT-QA}

PCDT-QA is an open-ended benchmark containing 890 questions (5 per guideline, 445 per split) covering both broad clinical topics (e.g., inclusion criteria) and specific factual queries (e.g., recommended pediatric dosages). Each question is paired with a reference answer grounded in the guideline text. Evaluation uses an LLM-as-a-judge pipeline: the model under evaluation generates a free-text answer, and GPT-4.1 compares it against the reference, producing a binary verdict (correct/incorrect). This accommodates the open-ended nature of clinical responses, where correct answers may differ substantially in wording from the reference.

\section{Experimental Setup}
\label{sec:setup}

\subsection{Continual Pre-Training}

We perform continual pre-training (CPT) on Qwen2.5-14B-Instruct~\cite{qwen25} using a standard causal language modeling objective over the full synthetic corpus. Training runs on a TPU v6-64 pod using JAX. We train for 3 epochs with a cosine learning rate schedule (peak $5 \times 10^{-5}$, minimum 10\% of peak) and AdamW ($\beta_1{=}0.9$, $\beta_2{=}0.95$). The effective batch size is 32. Sequences are truncated to 4{,}096 tokens; all tokens contribute to the loss. We use a warmup ratio of $1/3$ of total steps.

\subsection{Reinforcement Learning with GRPO}

After CPT, we apply GRPO~\cite{deepseek-math} using the HealthBench-BR train split. The task requires the model to assess factual clinical assertions and produce a justified true/false verdict, making it suitable for reinforcement learning with verifiable rewards (RLVR).

Following recent evidence that parameter-efficient methods are effective for RLVR~\cite{schulman2025lora,yin2025peftrlvr}, we train GRPO with LoRA ($r{=}32$, $\alpha{=}64$) applied to all linear layers for 5 epochs. We generate 16 completions per prompt with a maximum length of 512 tokens, use an effective batch size of 128, and set the KL coefficient $\beta$ to 0, relying on clipping ($\epsilon{=}0.2$, $\epsilon_{\mathrm{high}}{=}0.28$) to bound updates. The reward is 1.0 for correct answers with at least 50 words of reasoning and 0.0 otherwise, discouraging bare-label reward hacking while encouraging explicit clinical justification.

\subsection{Supervised Fine-Tuning}

As an alternative to GRPO, we also evaluate supervised fine-tuning (SFT) on guideline-grounded QA pairs. For each HealthBench-BR train-split question, we prompt GPT-5.2 with the source guideline in context to produce a reasoned \textit{Verdadeiro}/\textit{Falso} answer, yielding 890 prompt--answer pairs; no test-split assertions are used. We fine-tune with AdamW (lr $3 \times 10^{-5}$), effective batch size 64, masking the loss over user tokens so it is computed only on assistant tokens.

\section{Results}






\subsection{Main Results}

\begin{table*}[!htb]
\centering
\caption{Main results on HealthBench-BR (true/false accuracy) and PCDT-QA (open-ended accuracy via LLM judge). Train and Test refer to the guideline split (Section~\ref{sec:benchmarks}). All values in \%. Baseline is Qwen2.5-14B-Instruct without domain-specific training.}
\label{tab:main-results}
\begin{tabularx}{\textwidth}{X r@{\hskip 8pt}r @{\hskip 24pt} r@{\hskip 6pt}r}
\toprule
 & \multicolumn{2}{c}{\textbf{HealthBench-BR}} & \multicolumn{2}{c}{\textbf{PCDT-QA}} \\
\cmidrule(lr){2-3} \cmidrule(lr){4-5}
\textbf{Model} & \textbf{Train} & \textbf{Test} & \textbf{Train} & \textbf{Test} \\
\midrule
\multicolumn{5}{l}{\textit{Ours}} \\
\midrule
Baseline (no training)        & 61.1 & 59.4 & 30.8 & 27.9 \\
\quad + RL only               & 68.7 & 64.2 & 31.5 & 29.4 \\
\quad + SFT only              & 74.0 & 62.1 & 32.8 & 29.4 \\
\quad + CPT                   & 67.1 & 69.6 & 73.5 & 66.3 \\
\quad + CPT + SFT             & 83.5 & 70.4 & 64.9 & 57.8 \\
\quad + CPT + RL              & 85.5 & 75.6 & 70.6 & 65.4 \\
\quad + CPT (4 generators)     & 70.9 & 71.1 & 91.0 & 86.3 \\
\quad + CPT (4 generators) + SFT & 87.0 & 76.4 & 87.6 & 81.1 \\
\quad + CPT (4 generators) + RL & 90.4 & 83.9 & 88.8 & 85.4 \\
\midrule
\multicolumn{5}{l}{\textit{Frontier models}} \\
\midrule
GPT-4.1                 & 73.1 & 73.3 & 70.1 & 70.3 \\
GPT-5.2 (instant)       & 72.5 & 76.3 & 69.2 & 71.7 \\
GPT-5.2 (high)          & 78.9 & 78.5 & 74.6 & 78.2 \\
Claude Sonnet 4.6       & 77.2 & 77.6 & 67.0 & 70.3 \\
Gemini 3.1 Pro          & 77.8 & 79.3 & 74.8 & 80.0 \\
\midrule
\multicolumn{5}{l}{\textit{RAG}} \\
\midrule
Google AI Overview      & 72.4 & 70.5 & 77.1 & 77.3 \\
Baseline + RAG (BM25) & 88.4 & 90.0 & 56.0 & 58.2 \\
Baseline + RAG (embeddings) & 86.5 & 87.9 & 83.6 & 86.7 \\
CPT + RAG (BM25)             & 87.4 & 89.0 & 79.8 & 78.9 \\
CPT + RAG (embeddings)       & 86.1 & 85.2 & 91.2 & 92.1 \\
CPT (4 gen) + RAG (BM25)           & 74.2 & 73.9 & 92.1 & 88.3 \\
CPT (4 gen) + RAG (embeddings)     & 74.0 & 71.1 & 92.1 & 94.8 \\
CPT (4 gen) + RL + RAG (BM25)      & 94.9 & 95.3 & 90.6 & 88.8 \\
CPT (4 gen) + RL + RAG (embeddings) & 94.2 & 94.0 & 92.6 & 93.3 \\
\bottomrule
\end{tabularx}
\end{table*}

Table~\ref{tab:main-results} summarizes performance across all configurations and baselines. The base Qwen2.5-14B-Instruct model, without any domain-specific training, achieves 59.4\% on HealthBench-BR and 27.9\% on PCDT-QA (test split), suggesting that guideline-specific knowledge is largely absent from the model's parameters.

Continual pre-training with a single generator raises test performance to 69.6\% on HealthBench-BR and 66.3\% on PCDT-QA---already approaching GPT-4.1 on both benchmarks. Scaling to four generators further improves PCDT-QA to 86.3\%, substantially surpassing all frontier models, while HealthBench-BR reaches 71.1\%. The disproportionate gain on PCDT-QA suggests that generator diversity primarily benefits free-form knowledge recall, where exposure to varied phrasings and explanatory styles translates directly into richer answers.

Following CPT with GRPO yields the strongest HealthBench-BR results: 83.9\% on the test split with four generators, a 12.8 percentage point gain over CPT alone. The effect on PCDT-QA is more nuanced: GRPO produces a small decrease from 86.3\% to 85.4\%, likely because the reward objective optimizes specifically for true/false clinical verification rather than open-ended generation. Despite this marginal trade-off, the final model remains above all frontier baselines on both benchmarks.

Among frontier models, Gemini 3.1 Pro is the strongest baseline at 79.3\% on HealthBench-BR and 80.0\% on PCDT-QA, followed closely by GPT-5.2 (high). Our best configuration outperforms Gemini 3.1 Pro by 4.6 and 5.4 percentage points, respectively, despite being a 14B-parameter open model. Notably, even our CPT-only configuration (4 generators, no GRPO) already surpasses all frontier models on PCDT-QA by over 6 points, suggesting that continual pre-training on guideline-specific synthetic data is highly effective for knowledge absorption in this domain.

We also compare against Google AI Overview,\footnote{\url{https://blog.google/products/search/generative-ai-google-search-may-2024}} which generates answers grounded in web search results---effectively a retrieval-augmented generation (RAG) system---and is arguably one of the most widely deployed sources of AI-generated clinical information, given its integration into Google Search. For evaluation, we submit each benchmark question as a search query and extract the generated summary when available. Notably, AI Overview does not produce a response for every query: it abstained on 11.7\% of HealthBench-BR and 4.9\% of PCDT-QA questions, and the reported accuracy considers only questions where a response was generated, giving it an inherent advantage. Even under this favorable comparison, our best configuration outperforms AI Overview by 13.4 points on HealthBench-BR and 8.1 points on PCDT-QA, demonstrating that targeted knowledge injection into a 14B model can surpass retrieval-augmented access to the open web.

An important dimension beyond accuracy is sycophancy---the tendency to agree with the user regardless of factual correctness. On HealthBench-BR, a well-calibrated model should predict ``True'' for approximately 50\% of questions, since the dataset is perfectly balanced. The base Qwen2.5-14B-Instruct is well calibrated at 49.9\%, but CPT shifts this to 58.9\%, suggesting that continual pre-training on synthetic data introduces a bias toward agreement, possibly because the training corpus tends toward affirmative framing. GRPO effectively corrects this: our best configuration (CPT + RL, 4 generators) returns to 53.0\%, indicating that the reinforcement learning stage, by requiring the model to justify its verdict, teaches it to disagree when the evidence warrants it. Among frontier models, Claude Sonnet 4.6 (49.2\%) and GPT-5.2 high (48.7\%) are well calibrated, while Gemini 3.1 Pro exhibits a moderate agreement bias at 63.3\%---noteworthy given that it is the strongest frontier baseline on both benchmarks. The most concerning case is Google AI Overview, which predicts ``True'' 69.6\% of the time, making it the most sycophantic system evaluated. Given its reach as the default AI-generated response in Google Search, this tendency to confirm clinical assertions---including false ones about dosages, contraindications, and monitoring intervals---raises concerns for users seeking clinical information without professional guidance.

We complement parametric evaluation with two RAG variants: BM25 and dense retrieval using \texttt{text-embedding-3-small} embeddings\footnote{\url{https://developers.openai.com/api/docs/models/text-embedding-3-small}}, both retrieving the top-$k{=}10$ chunks (2{,}000 characters, 200-character overlap) from the full guideline corpus. Retriever choice interacts with task format: BM25 dominates HealthBench-BR (95.3 vs.\ 94.0 with our best model), where verifying dosages and intervals rewards lexical matching, while embeddings dominate PCDT-QA (93.3 vs.\ 88.8), where semantic retrieval surfaces broader evidence. Notably, CPT (4 gen) + RAG underperforms Baseline + RAG on HealthBench-BR (73.9 vs.\ 90.0), suggesting that a domain-adapted model partially overrides retrieved evidence---an effect GRPO reverses (95.3), as the verdict reward trains the model to ground its answer in the retrieved context.

\subsection{Effect of Generation Formats}

\begin{table*}[t]
\centering
\caption{Effect of generation format and generator diversity on CPT performance (test split). HB = HealthBench-BR, PCDT = PCDT-QA. ``1 generator'' uses GPT-4.1-mini only; ``4 generators'' uses all four models described in Section~\ref{sec:synthesis}.}
\label{tab:ablation-augtype}
\begin{tabularx}{\textwidth}{X r@{\hskip 6pt}r@{\hskip 6pt}r @{\hskip 24pt} r@{\hskip 6pt}r@{\hskip 6pt}r}
\toprule
 & \multicolumn{3}{c}{\textbf{1 generator}} & \multicolumn{3}{c}{\textbf{4 generators}} \\
\cmidrule(lr){2-4} \cmidrule(lr){5-7}
\textbf{Generation format} & \textbf{\#} & \textbf{HB} & \textbf{PCDT} & \textbf{\#} & \textbf{HB} & \textbf{PCDT} \\
\midrule
Rephrase              & 1{,}780  & 62.8 & 52.4 &  7{,}120 & 63.6 & 73.0 \\
Wiki                  & 1{,}780  & 64.2 & 60.9 &  7{,}120 & 69.7 & 76.9 \\
Rephrase + QA         & 2{,}670  & 63.7 & 56.6 & 10{,}680 & 66.3 & 75.3 \\
Wiki + QA             & 2{,}670  & 67.0 & 62.2 & 10{,}680 & 68.2 & 76.0 \\
Wiki + Rephrase       & 3{,}560  & 63.0 & 65.4 & 14{,}240 & 68.4 & 83.8 \\
Rephrase + Wiki + QA  & 4{,}450  & 69.6 & 66.3 & 17{,}800 & 71.1 & 86.3 \\
\bottomrule
\end{tabularx}
\end{table*}

Table~\ref{tab:ablation-augtype} isolates the contribution of each augmentation format under both single- and multi-generator settings. Across all configurations, wiki-style articles consistently outperform rephrases---for instance, with a single generator, wiki alone reaches 64.2\% on HealthBench-BR and 60.9\% on PCDT-QA, compared to 62.8\% and 52.4\% for rephrase. This advantage likely reflects the expository nature of the wiki format, which restructures protocol content into self-contained explanations that more closely resemble the reasoning required at evaluation time.

Combining formats is consistently beneficial: each additional format improves at least one benchmark, and the full combination (rephrase + wiki + QA) yields the best results in every setting. With four generators, adding QA to the wiki + rephrase combination raises HealthBench-BR from 68.4\% to 71.1\% (+2.7 points) and PCDT-QA from 83.8\% to 86.3\% (+2.5 points), confirming that direct question--answer supervision provides a complementary learning signal that document-style formats alone do not fully capture.

\subsection{Effect of the Generator}
\label{sec:ablation-generators}

\begin{table*}[t]
\centering
\caption{Effect of the generator model on CPT performance. Each model generates all three augmentation types (rephrase, wiki, QA) for 4{,}450 training examples.}
\label{tab:ablation-generators}
\begin{tabular}{l @{\hskip 18pt} r @{\hskip 18pt} r @{\hskip 18pt} r @{\hskip 12pt} r}
\toprule
\textbf{Generator model} & \textbf{Tokens} & \textbf{Avg.\ doc} & \textbf{HB (\%)} & \textbf{PCDT (\%)} \\
\midrule
Qwen2.5-14B (self-aug) & 7.1M  & 1{,}594 & 60.2 & 50.1 \\
GPT-OSS-20B            & 15.4M & 3{,}471 & 65.7 & 64.0 \\
GPT-5-nano             & 15.7M & 3{,}522 & 65.2 & 67.4 \\
GPT-4.1-mini           & 18.1M & 4{,}065 & 69.6 & 66.3 \\
Qwen3-235B             & 20.3M & 4{,}551 & 63.5 & 66.7 \\
\bottomrule
\end{tabular}
\end{table*}

Table~\ref{tab:ablation-generators} compares the five generator models in isolation, each producing all three generation formats. The clear outlier is self-generation with the base model itself (Qwen2.5-14B-Instruct), which yields the weakest results by a wide margin---60.2\% on HealthBench-BR and 50.1\% on PCDT-QA. We attribute this to output length: Qwen2.5-14B averages 1{,}594 tokens per example against 3{,}471--4{,}551 for the external generators, producing only 7.1M tokens total. Shorter outputs inevitably omit the specific dosages, monitoring intervals, and inclusion criteria that are central to both benchmarks.

Among the four external generators, no single model leads on both benchmarks: GPT-4.1-mini achieves the best HealthBench-BR score (69.6\%), while GPT-5-nano leads on PCDT-QA (67.4\%). Notably, Qwen3-235B does not dominate either benchmark despite producing the most tokens, suggesting diminishing returns from additional generator capacity.

This motivates combining all four external generators rather than relying on the strongest one. Generator and format diversity are complementary: with one generator, all three formats combined yield 66.3\% on PCDT-QA; scaling to four generators on the same combination reaches 86.3\% (Table~\ref{tab:ablation-augtype})---a 20.0 point gain that far exceeds any single-format improvement.

\subsection{Catastrophic Forgetting Analysis}

\begin{table}[!htb]
\centering
\caption{Comparison of catastrophic forgetting-mitigation strategies. All values are test accuracy in \%. HealthBench-BR and PCDT-QA are in-distribution benchmarks; Avg is the average over the five OOD benchmarks.}
\label{tab:cpt-lora}
\begin{tabular}{llcccccccc}
\toprule
 & & \multicolumn{2}{c}{\textbf{In-distribution}} & \multicolumn{6}{c}{\textbf{Out-of-distribution}} \\
\cmidrule(lr){3-4} \cmidrule(lr){5-10}
\textbf{Data} & \textbf{Method} & HB & PCDT & HealthQA & DrBode & MedPT & BLUEX & IFEval & Avg \\
\midrule
Baseline & -- & 59.4 & 27.9 & 67.8 & 65.3 & 56.4 & 75.0 & 79.7 & 68.8 \\
\midrule
\multirow{3}{*}{1 gen} & Full FT & 69.6 & 66.3 & 66.0 & 62.7 & 56.4 & 73.5 & 75.8 & 66.9 \\
 & LoRA & 67.2 & 65.2 & 64.1 & 56.6 & 53.8 & 71.1 & 68.6 & 62.8 \\
 & Replay & 63.8 & 65.6 & 67.0 & 64.5 & 55.9 & 72.7 & 77.1 & 67.4 \\
\midrule
\multirow{3}{*}{4 gen} & Full FT & 71.1 & 86.3 & 63.3 & 61.2 & 54.1 & 72.8 & 69.3 & 64.1 \\
 & LoRA & 70.3 & 77.1 & 60.8 & 58.3 & 52.1 & 68.9 & 65.6 & 61.1 \\
 & Replay & 69.2 & 80.0 & 65.7 & 59.2 & 54.0 & 73.2 & 69.3 & 64.3 \\
\bottomrule
\end{tabular}
\end{table}

Table~\ref{tab:cpt-lora} reports performance on five out-of-distribution (OOD) benchmarks alongside our two in-distribution evaluations, comparing two forgetting-mitigation strategies. The OOD set spans HealthQA-BR~\cite{daddario2025healthqabr}, DrBode~\cite{recogna2024drbode}, MedPT~\cite{farber2025medpt}, BLUEX~\cite{almeida2023bluex} (Brazilian university admissions, general PT knowledge), and IFEval~\cite{zhou2023ifeval} (instruction-following). We compare full fine-tuning, LoRA, and full FT with 50\% replay drawn from FineWeb~\cite{penedo2024fineweb} as an English-language proxy for the Qwen2.5 pre-training distribution.

Full fine-tuning (FT) yields moderate forgetting: $-1.9$ points on average with one generator and $-4.7$ with four. The larger drop tracks the larger in-distribution gain (PCDT-QA $66.3 \rightarrow 86.3$), consistent with the power-law trade-off in~\cite{kalajdzievski2024scaling}. IFEval is the most affected ($-4$ to $-10$ points), suggesting that prolonged optimization on document-style text partially erodes instruction-following alignment.

Counterintuitively, LoRA is dominated by full FT on both axes---more OOD forgetting ($-6.0$ vs.\ $-1.9$ with one generator; $-7.7$ vs.\ $-4.7$ with four) and lower in-distribution scores (PCDT-QA $77.1$ vs.\ $86.3$). This contrasts with Biderman et al.~\cite{biderman2024loraforgets}, who report that LoRA both learns and forgets less than full FT. The pattern is instead consistent with the \emph{intruder dimensions} account of Shuttleworth et al.~\cite{shuttleworth2024intruder}, in which LoRA solutions contain high-magnitude singular vectors approximately orthogonal to the base model---absent from full-FT solutions---that degrade unrelated tasks; this matches our largest gap being on IFEval ($-11.1$ and $-14.1$ vs.\ $-3.9$/$-10.4$), where these directions plausibly overlap with instruction-following parameters.

Mixing FineWeb at a 1:1 ratio helps mainly at smaller scale: with one generator, replay recovers the OOD average to $67.4$ ($-1.4$ vs.\ baseline; $+0.5$ over full FT) while retaining most of the in-distribution gain (HealthBench-BR $63.8$, PCDT-QA $65.6$). At four generators the benefit vanishes ($64.3$ vs.\ full FT's $64.1$), and the in-distribution cost shifts from HealthBench-BR ($\sim$6 points at one generator) to PCDT-QA ($80.0$ vs.\ $86.3$). Since FineWeb is predominantly English, part of the retention may reflect continued English exposure; a PT replay source is a natural follow-up, consistent with Bethune et al.~\cite{bethune2025scaling}.

Overall, full FT sits on the favorable side of the knowledge--retention frontier across both data scales; replay offers a small OOD-preservation benefit at lower training scale that diminishes with more training data, at a consistent in-distribution cost; LoRA is dominated on both axes. We adopt full FT for the main results.

\section{Ethical Considerations}


Our model is designed to support, not replace, clinical judgment: even at 83.9\% on HealthBench-BR, residual errors may involve consequential details such as dosages or contraindications. The released artifact is a research model, not a certified medical device, and should be used only under professional supervision. Since all training data comes from public Ministry of Health guidelines, privacy risks from clinical notes are avoided, though biases from the source documents and generator models may persist. Our sycophancy results further show that clinical AI systems should be evaluated not only for accuracy but also for their tendency to confirm false user assertions. Finally, open release of models trained on Brazilian protocols is a step toward equitable access for Portuguese-speaking clinicians and patients underserved by English-centric medical LLMs.

\section{Conclusion}

We presented a reproducible protocol for adapting an open LLM to the Brazilian Unified Health System, together with two protocol-grounded benchmarks---HealthBench-BR and PCDT-QA---that fill an evaluation gap for Portuguese clinical NLP. Expanding 178 official guidelines (5.4M tokens) into ${\sim}$70M tokens of synthetic data across three formats and four generators, then applying continual pre-training and GRPO to Qwen2.5-14B-Instruct, yields 83.9\% on HealthBench-BR and 85.4\% on PCDT-QA, outperforming GPT-5.2, Claude Sonnet 4.6, Gemini 3.1 Pro, and Google AI Overview's web-grounded RAG by 4.6 to 15.1 points despite being a 14B open model. Ablations show that generator and format diversity are complementary (+20 points on PCDT-QA), GRPO is critical for verdict-style recall (+12.8 points) and corrects post-CPT sycophancy from 58.9\% to 53.0\%, and full FT dominates LoRA on both knowledge gain and OOD retention. Limitations include evaluation grounded in guideline recall rather than real clinical encounters and the absence of prospective validation with professionals. We release all datasets, benchmarks, and model weights to support reproducible research on clinical NLP for Brazilian Portuguese.

%
%
%
\bibliographystyle{splncs04}
\bibliography{mybibliography}

\end{document}